\documentclass[10pt,journal,compsoc]{IEEEtran}

\usepackage[utf8]{inputenc}
\usepackage{amssymb,amsfonts,amsmath}
\usepackage{algorithm,algpseudocode,amsthm,mathrsfs,subcaption}
\usepackage{lipsum}
\usepackage{hyperref}
\hypersetup{colorlinks,
}
\usepackage{graphicx}
\usepackage{hyperref}
\usepackage{array}
\usepackage{stfloats}
\usepackage{booktabs}
\usepackage{csquotes}
\usepackage{subcaption}

\newcommand{\Pbb}{\text{I\kern-0.15em P}}

\begin{document}

\title{Universal Adversarial Perturbations: Efficiency on a small image dataset}

\author{Waris Radji*\thanks{*Student in the introductory research program of Bordeaux Institute of Technology - ENSEIRB-MATMECA School of Engineering.}}

\IEEEtitleabstractindextext{%
}

% make the title area
\maketitle

\IEEEdisplaynontitleabstractindextext
\IEEEpeerreviewmaketitle

\section*{\hfil Abstract \hfil}
{\itshape

Although neural networks perform very well on the image classification task, they are still vulnerable to adversarial perturbations that can fool a neural network without visibly changing an input image. A paper has shown the existence of Universal Adversarial Perturbations which when added to any image will fool the neural network with a very high probability. In this paper we will try to reproduce the experience of the Universal Adversarial Perturbations paper, but on a smaller neural network architecture and training set, in order to be able to study the efficiency of the computed perturbation.
}
\section{Introduction}
\label{sec:introduction}

As part of my engineering school's introductory research program, I aim to reproduce a scientific experiment. I chose to reproduce the \textbf{universal adversarial perturbations} paper experiments introduced by Seyed-Mohsen Moosavi-Dezfooli \textit{et al} accepted at IEEE Conference on Computer Vision and Pattern Recognition (CVPR) in 2017\cite{1610.08401}. 

\begin{figure}[h]

\centering
\includegraphics[width=0.4\textwidth]{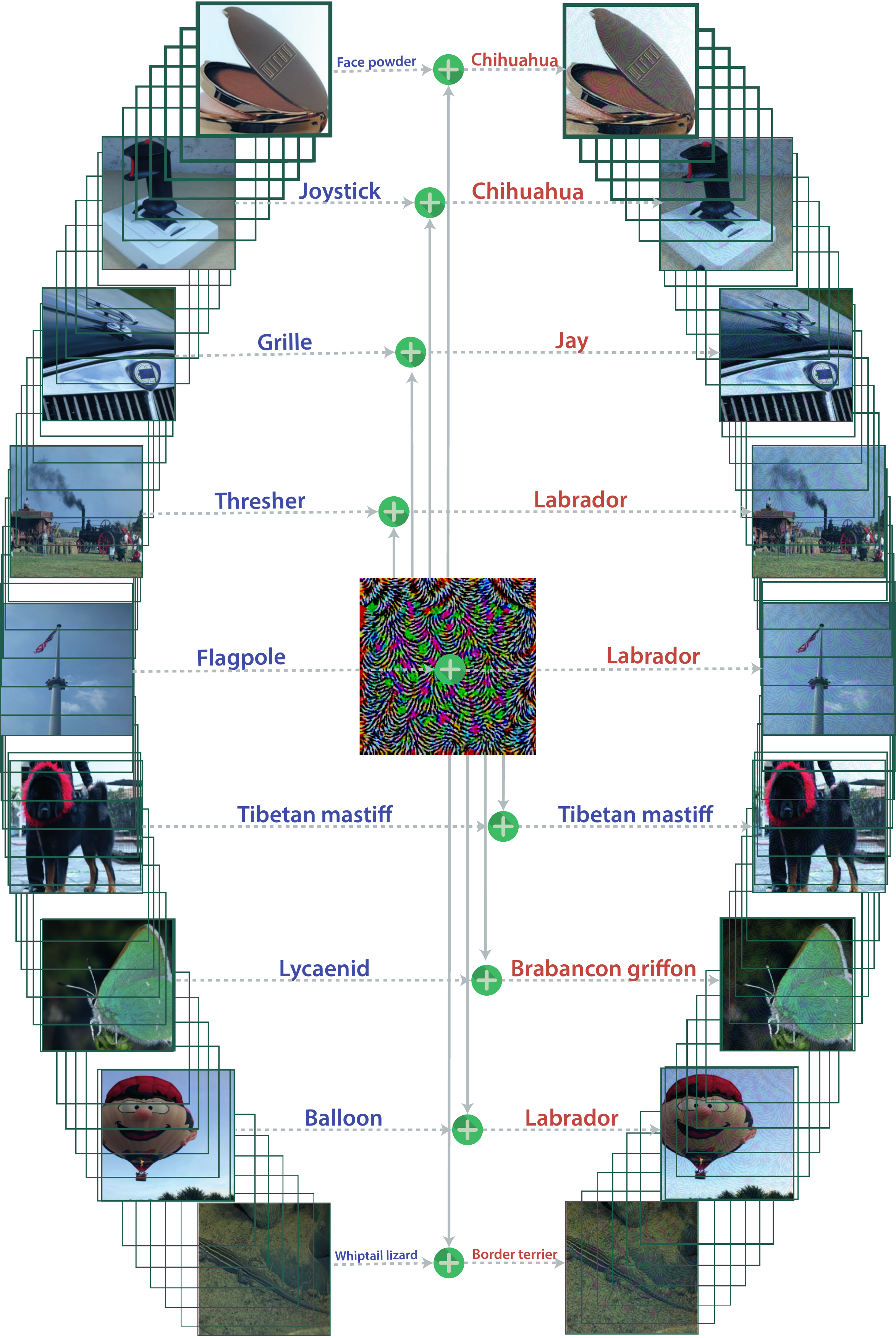}
\caption{This figure is taken from the reference article. When added to a natural image, a universal perturbation image causes the image to be misclassified by the deep neural network with high probability. Left images: Original natural images. The labels are shown on top of each arrow. Central image: Universal perturbation. Right images: Perturbed images.}
\label{img:univ} 
\end{figure}

In image classification, an adversarial perturbation is a vector that when added to an image cause the network output to change drastically with making quasi-imperceptible changes to the input image. Universal adversarial perturbations (UAPs) correspond to a single vector that will fool the neural network on any image with very high probability (see Fig.\ref{img:univ}). These perturbations have many applications in the real world, for example in CAPTCHA tests when a user has to associate an image with a label or in steganography to hide information in images\cite{UDDIN2020146}. They can also be used for hostile purposes, knowing that they are very difficult to detect with the naked eye.

The \textbf{reference paper} \cite{1610.08401} has already demonstrated the existence and efficiency of UAPs, on large neural network architecture (CaffeNet, VGG-F, VGG-16, VGG-19, GoogLeNet, and ResNet-152) and with the very large image dataset of ImageNet Large Scale Visual Recognition Challenge 2012 (ILSVRC2012). In this paper we will try to reproduce some of their results, with VGG-11 \cite{1409.1556} neural network architecture which is smaller than those studied in the paper and on the Visual Object Classes Challenge 2012 (VOC2012) \cite{VOC2012} dataset which is 1,000 times smaller than ILSVRC2012: 

\begin{itemize}
    \item We will train a neural network;
    \item We will compute the UAP with parameters quasi-similar to the paper one;
    \item We will observe dominant labels founded by the algorithm; 
    \item We will compare the computed UAP to other perturbations.
\end{itemize}

All the code that allowed me to achieve my experiments can be found on \href{https://github.com/riiswa/universal}{Github}.

\section{Universal adversarial perturbations computation}
\label{sec:univ}

Before conclusions can be drawn from the UAPs, the method by which they are estimated must be studied.

\subsection{Notations and terminology}

The important mathematical notations used in the reference paper will be explained in this section.

\begin{itemize}
    \item $\mu$: the distribution of images in $\mathbb{R}^d$.
    \item $x \sim \mu$: an non-perturbed input image.
    \item $v$: the universal perturbation vector (UAP).
    \item $X = \{x_1, \dots, x_n\}$: a set of images sampled from the distribution $\mu$. Input of the algorithm \ref{alg:finding_universal_perturbations}.
    \item $X_v$: $X$ for which each image is summed with $v$. The perturbed set.
    \item $\hat{k}$: a classification function that outputs for each image $x \in \mathbb{R}^d$ an estimated label $\hat{k}(x)$.
    \item $\Delta v_i$: minimal perturbation that sends the current perturbed point $x_i +v$ to the decision boundary of the classifier.
    \item \textit{fooling rate}: the proportion of total perturbed images ($x + v$) in a dataset for which $\hat{k}(x + v) \neq \hat{k}(x)$
    \item $\xi$: parameter of the algorithm \ref{alg:finding_universal_perturbations} that controls the magnitude of $v$.
    \item $\delta$: parameter algorithm \ref{alg:finding_universal_perturbations} that quantifies the desired fooling rate for all images in $X$.
    
\end{itemize}

We seek a vector $v$ such that

\begin{equation*}
    \hat{k}(x + v) \neq \hat{k}(x) \text{ for "most" } x \sim \mu
\end{equation*}
In simple words, the UAP that maximizing the fooling rate. There are two constraints to find $v$:

\begin{enumerate}
    \item $\| v \|_p \leq \xi,$
    \item $\underset{x \sim \mu}{\Pbb} \left( \hat{k} (x+v) \neq \hat{k} (x) \right) \geq 1 - \delta.$
\end{enumerate}
The first constraint ensures the quasi-imperceptible nature of the UAP, and the second allows the Algorithm \ref{alg:finding_universal_perturbations} to iterate until the desired fooling rate is reached.

\subsection{Algorithm}
\label{sec:algo}

\begin{figure}[h]

\centering
\includegraphics[width=0.35\textwidth]{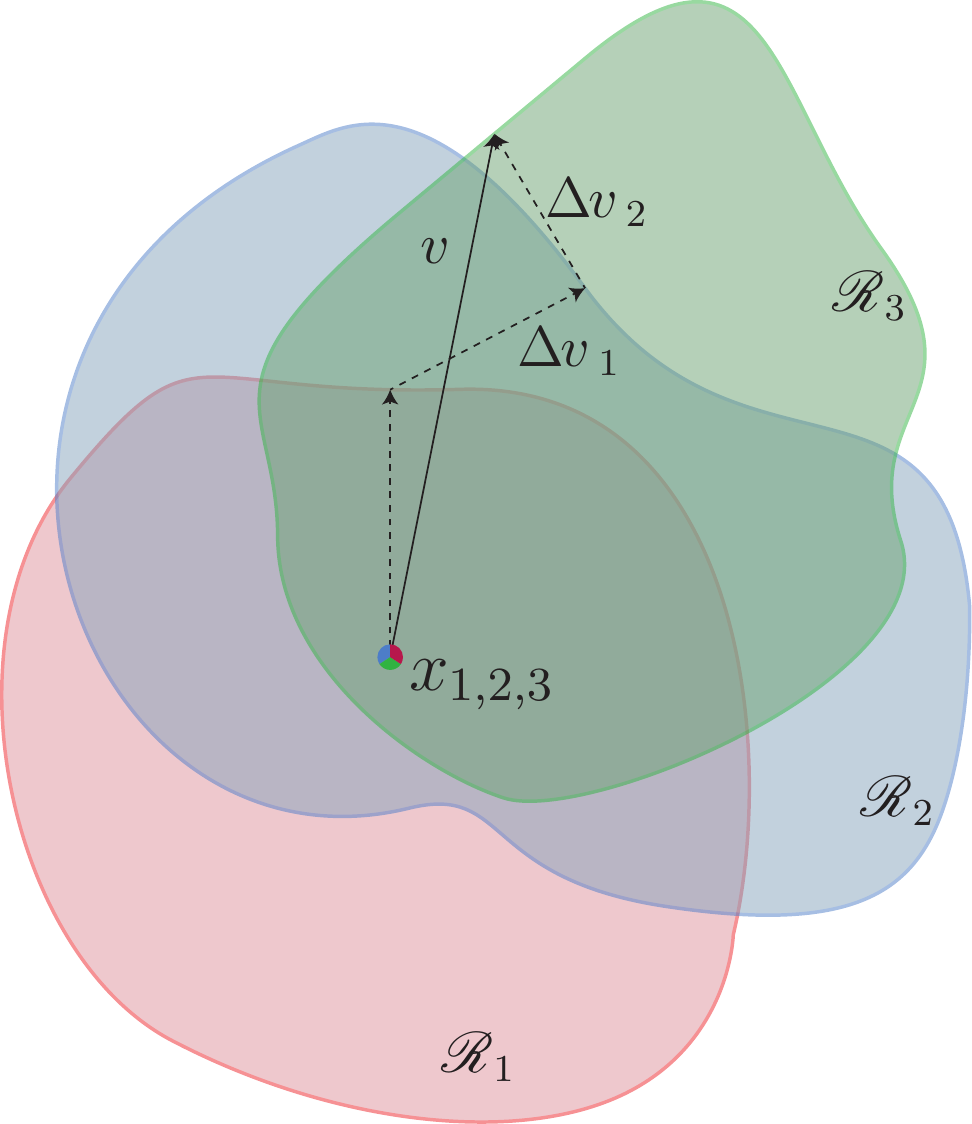}
\caption{This figure is taken from the reference article. Schematic representation of the algorithm used to compute universal perturbations. In this illustration, data points $x_1, x_2$ and $x_3$ are super-imposed, and the classification regions $\mathscr{R}_i$ (i.e., regions of constant estimated label) are shown in different colors. Our algorithm proceeds by aggregating sequentially the minimal perturbations sending the current perturbed points $x_i+v$ outside of the corresponding classification region $\mathscr{R}_i$.}
\label{img:illust_alg} 
\end{figure}

Until the desired fooling rate is not reached (it is also possible to define a maximum number of iterations) the algorithm takes each well classified data point in $X$ and computes the minimal perturbation $\Delta v_i$ that sends the current perturbed point $x_i +v$ to the decision boundary of the classifier as illustrated in Fig. \ref{img:illust_alg}.  $\Delta v_i$ is computed by solving the following optimisation problem :

\begin{equation}
\label{eq:deepfool}
\Delta v_i \gets \arg\min_{r} \| r \|_2 \text{ s.t. } \hat{k} (x_i + v + r) \neq \hat{k} (x_i).
\end{equation}
The resolution of Equation \ref{eq:deepfool} is allowed by the \textit{DeepFool} algorithm \cite{1511.04599} of the same authors which computes the minimal perturbation of a single image. To control the norm of the vector $v$ according to the parameter $\xi$, the updated perturbation vector is projected on the $\ell_p$ ball of radius $\xi$ and centered at 0. The projection operator $\mathcal{P}_{p, \xi}$ is defined as follows:

\begin{equation*}
    \mathcal{P}_{p, \xi} (v) = \arg\min_{v'} \| v - v' \|_2 \text{ subject to } \| v' \|_p \leq \xi.
\end{equation*}
The updated vector $v$ is given by $v \gets \mathcal{P}_{p, \xi} (v + \Delta v_i)$.

The detailed algorithm is provided in Algoritm \ref{alg:finding_universal_perturbations}. Based on the \href{https://github.com/LTS4/universal}{code} provided by the authors, I was able to adapt this algorithm to \textit{Pytorch} for my experiment.

\begin{algorithm}[ht]

\caption{Computation of universal perturbations.}
\begin{algorithmic}[1]
\State \textbf{input:} Data points $X$, classifier $\hat{k}$, desired $\ell_p$ norm of the perturbation $\xi$, desired accuracy on perturbed samples $\delta$.
\State \textbf{output:} Universal perturbation vector $v$.
\State Initialize $v \gets 0$.
\While{$\text{Err} (X_v) \leq 1-\delta$}
\For{each datapoint $x_i \in X$}
\If{$\hat{k}(x_i+v) = \hat{k} (x_i)$}
\State Compute the \textit{minimal} perturbation that sends $x_i+v$ to the decision boundary:
\begin{align*}
\quad \quad \Delta v_i \gets \arg\min_{r} \| r \|_2 \text{ s.t. } \hat{k} (x_i + v + r) \neq \hat{k} (x_i).
\end{align*}
\State Update the perturbation: 
$$v \gets \mathcal{P}_{p, \xi} (v+\Delta v_i).$$
\EndIf
\EndFor
\EndWhile
\end{algorithmic}
\label{alg:finding_universal_perturbations}
\end{algorithm}

\section{Universal Perturbation of VGG-11 neural network architecture}

In this section we will reproduce the experiment of the reference paper \cite{1610.08401} which consists in computing an UAP, based on the VGG-11 neural network architecture \cite{1409.1556} and on the VOC212 dataset \cite{VOC2012}.

\subsection{The VOC2012 dataset}

Not having the computational resources of the authors of the article, I chose to use the VOC2012 dataset which is quite similar to the ILSVRC2012 dataset which is used in the article, but much smaller. Table \ref{table1} presents a summary of the comparison of the two datasets. The twenty VOC2012 classes can be seen in detail in Table \ref{tab:classes}.

\begin{table*}[t]

\caption{Comparison between VOC2012 and ILSVRC2012}
\label{table1}

\begin{tabular}{@{}lcc@{}}
\toprule
\textbf{Criterion}                       & \textbf{VOC2012}                                                                                                                                         & \textbf{ILSVRC2012}                                                                                                                                                    \\ \midrule
\multicolumn{1}{|l|}{Number of images}    & \multicolumn{1}{c|}{Both 5,000 images in training and validation set.}                                                                                     & \multicolumn{1}{c|}{1.28 million training images, and 50 thousand validation image.}                                                                                     \\ \midrule
\multicolumn{1}{|l|}{Number of classes}  & \multicolumn{1}{c|}{20}                                                                                                                                  & \multicolumn{1}{c|}{1000}                                                                                                                                              \\ \midrule
\multicolumn{1}{|l|}{Variety of classes} & \multicolumn{2}{c|}{\begin{tabular}[c]{@{}c@{}} Both datasets contains a variety of objects and living things. ILSVRC2012 divides its data more than VOC2012, for \\example the cat class of VOC2012 is divided into cat breeds in ILSVRC2012 (Egyptian, Persian, tiger, Siamese, etc... ).\end{tabular}} \\ \midrule
\multicolumn{1}{|l|}{Image size}         & \multicolumn{2}{c|}{About 300 $\times$ 300}                                                                                                                                                                                                                                                                           \\ \midrule
\multicolumn{3}{c}{Both dataset are similar with respect to the average number of object instances and the amount of clutter per image.}                                                                                                                                                                                                                                      \\ \bottomrule
\end{tabular}
\end{table*}

\begin{table}[h]
    \centering
    \begin{tabular}{c|l}
        Person & Person \\
         Animal & bird, cat, cow, dog, horse, sheep \\
         Vehicle & aeroplane, bicycle, boat, bus, car, motorbike, train \\
         Indoor & bottle, chair, dining table, potted plant, sofa, tv/monitor
    \end{tabular}
    \caption{The twenty object classes of VOC2012}
    \label{tab:classes}
\end{table}

\subsection{Training VGG-11 neural network}

For the architecture of the neural network, I chose VGG-11 which is quite similar to some of the architectures studied in the reference paper (VGG-16, VGG-19), but which has fewer convolutional layers, thus fewer parameters. This architecture is have a higher classification error rate.

Instead of taking a randomly initialized VGG-11 model, \textit{Torchvision} offers pre-trained models. The \textit{Torchvision} models have been trained on the ImageNet dataset. In our case VOC2012 containing only 20 classes. To adapt the pre-trained model we define a new final layer which has an input size equal to the layer we are replacing (4096), but with an output of 20 classes. This new layer is randomly initialized and is the only part of our model that does not have pre-trained parameters.

To load the data into the model, simply follow the \href{https://pytorch.org/vision/stable/models.html}{Torchvision} instructions :
\begin{displayquote}
All pre-trained models expect input images normalized in the same way, i.e. mini-batches of 3-channel RGB images of shape (3 x H x W), where H and W are expected to be at least 224. The images have to be loaded in to a range of [0, 1] and then normalized using \texttt{mean = [0.485, 0.456, 0.406]} and \texttt{std = [0.229, 0.224, 0.225]}.
\end{displayquote}

We finally reach an accuracy score of 75\% on the validation dataset. Considering that VOC2012 is also a dataset that is used for image segmentation, which means that an image can have several labels (e.g. a person walking his dog), which biases the model. For ease of use, the score is based on only one of the true labels of the image.

\subsection{Computation of the UAP}

Now that we have our classifier, we can compute the UAP, by using the algorithm \ref{alg:finding_universal_perturbations} with the parameters $\xi = 2000$ for the norm $\ell_2$, and a desired fooling rate $\delta$ of 0.80. The implementation of the algorithm has been slightly modified to be adapted to \textit{Pytorch} on GPU and to reduce memory usage.

The image set $X$ was chosen by taking 50 images per class from the validation dataset, i.e. a total of 1000 images. Due to the small volume of data compared to the reference paper (10,000 images), it was not possible to reach the desired fooling rate of 80\%, but a fooling rate of 56\% was achieved, which is consistent with an experiment in the paper that showed that the fooling rate was strongly dependent on the size of $X$.

\begin{figure}[h]
\centering
\includegraphics[width=0.35\textwidth]{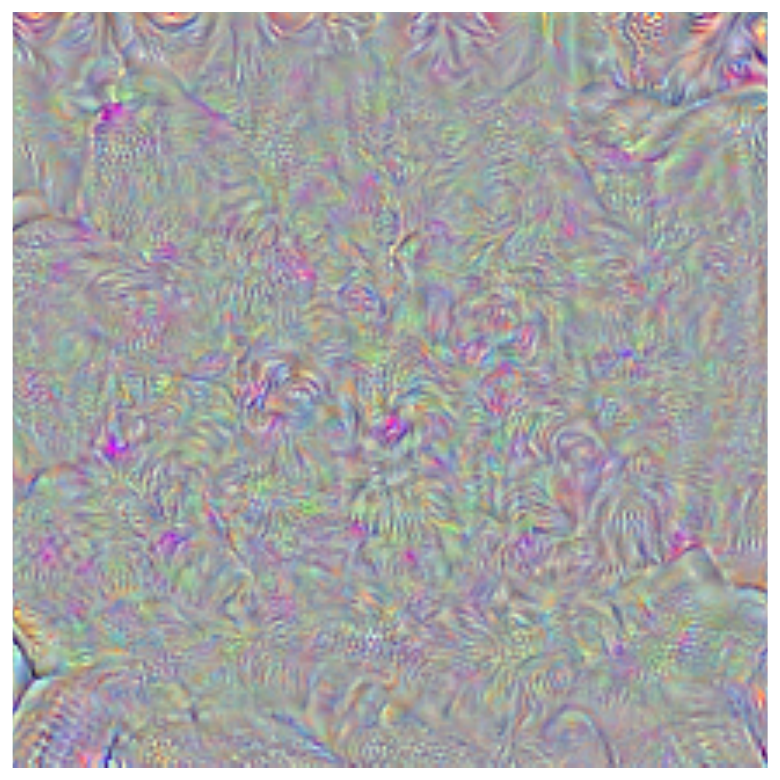}
\caption{UAP computed for VGG-11 architecture.}
\label{img:mypert} 
\end{figure}

It is now possible to visualise the UAP that corresponds to the VGG-11 architecture in Figure \ref{img:mypert}. The perturbation obtained is very different from those obtained in the reference paper (see Figure \ref{vgg}).

\begin{figure}[h]
\centering
\includegraphics[width=0.47\textwidth]{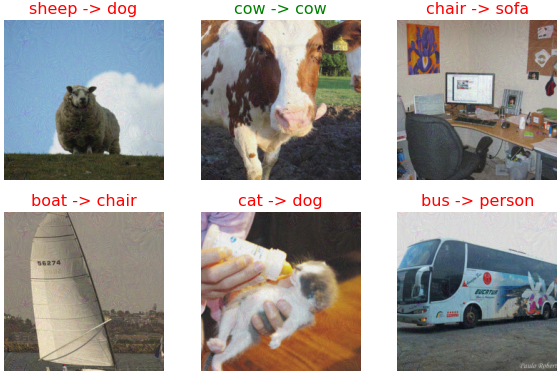}
\caption{Examples of perturbed images and their corresponding labels tranformation.}
\label{img:perturbed_images} 
\end{figure}

Figure \ref{img:perturbed_images} shows the results of adding the UAP on images from the VOC2012 validation set. The UAP is quasi-imperceptible but succeeds in fooling the neural network on a large number of images.

\section{Dominant labels}

\begin{figure}[h]
\centering
\includegraphics[width=0.43\textwidth]{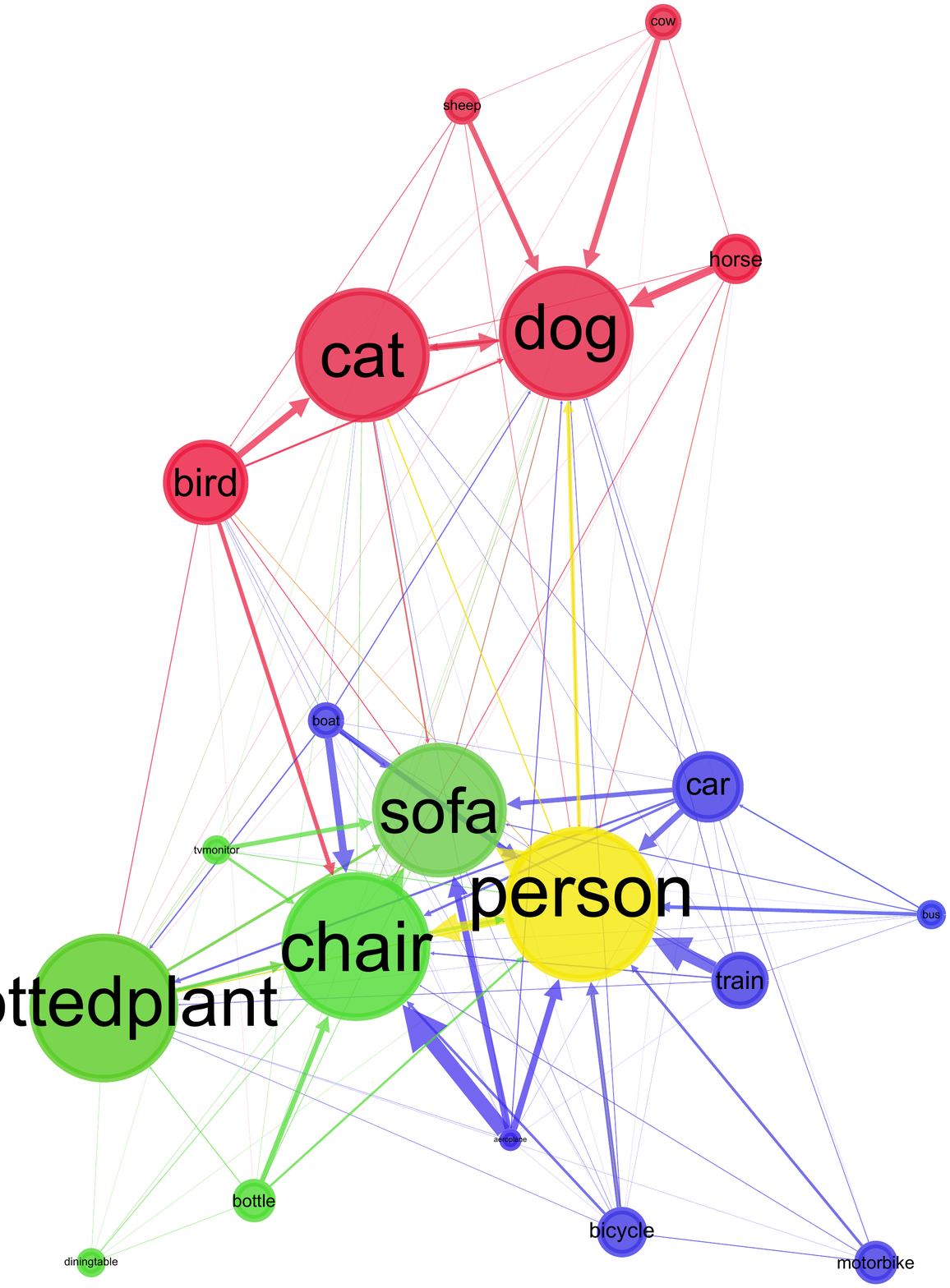}
\caption{Representation of the graph $G = (V, E)$, where the vertices are the set of labels, and weighted directed edges $i \rightarrow j$ indicate the number of label $i$ are fooled into label $j$ when applying the universal perturbation.}
\label{gephi} 
\end{figure}

To better see the effect of UAP on natural images, we can view the label distribution of VOC2012. We build a directed graph $G = (V, E)$, whose vertices denote the labels, and weighted directed edges $e = (i \leftarrow j)$ indicate the number of label $i$ are fooled into label $j$ when applying the universal perturbation. The color of the vertice represents the category group (see Table \ref{tab:classes}) to which it belongs and the size of the node represents its indegree. The graph in Figure \ref{gephi} was generated with the \textit{Gephi} tool and the vertices were placed automatically with the \textit{ForceAltas2} \cite{gephi} algorithm.

By observing the topoly of the graph obtained, we can see that the \textit{ForceAltas2} algorithm has automatically (in an unsupervised way) grouped the labels belonging to the same category, which results in strong links between these labels. Labels belonging to the same category tend to have a relationship in terms of which label the neural network will be fooled into (e.g. cat and dog). It is possible to see some labels stand out by their sizes. These labels are called in the reference article the \textit{dominant labels}, and the perturbations are mostly formed from these labels. The hypothesis in the reference article to explain this, is that these dominant labels occupy large regions in the image space, and therefore represent good candidate labels for fooling most natural images.

\section{Comparing with other perturbations}

\begin{figure}[h]
     \centering
     \begin{subfigure}[b]{0.2\textwidth}
         \centering
         \caption{VGG-16}
         \label{vgg16}
         \includegraphics[width=\textwidth]{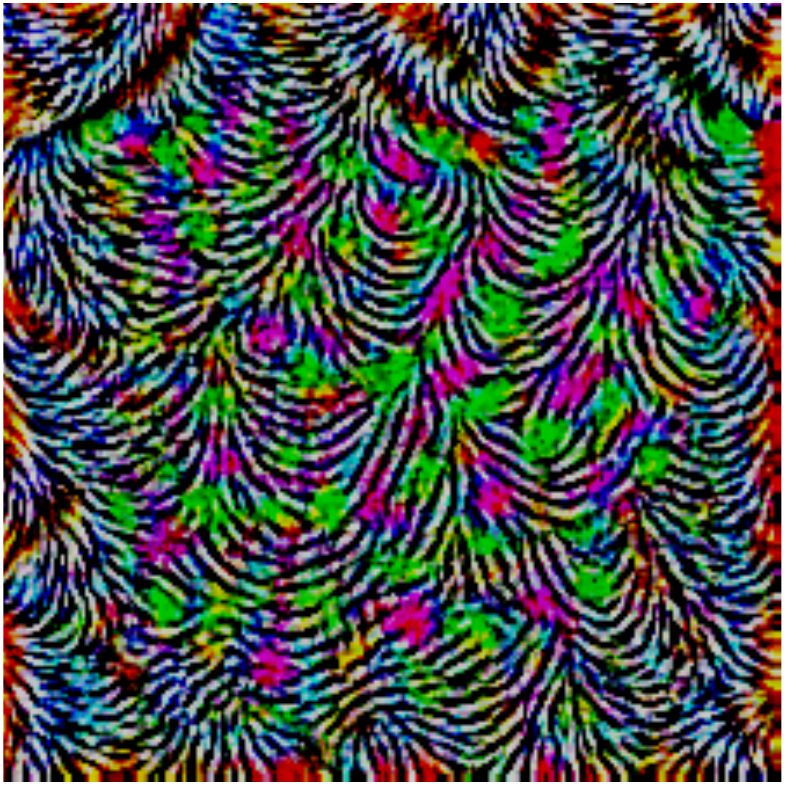}
     \end{subfigure}
     \begin{subfigure}[b]{0.2\textwidth}
         \centering
         \caption{VGG-19}
         \label{vgg19}
         \includegraphics[width=\textwidth]{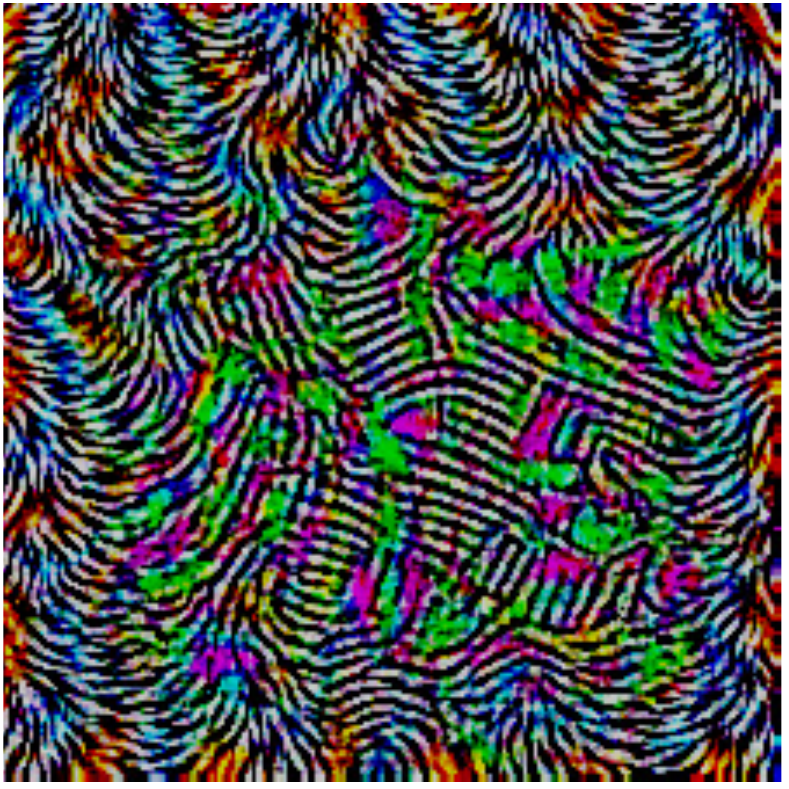}
     \end{subfigure}
     \caption{UAPs from the reference paper}
     \label{vgg}
\end{figure}

In this section we will try to understand the unique characteristics of universal pertubations by comparing our perturbations with other ones:
\begin{itemize}
    \item A random perturbation;
    \item UAPs computed by the Algorithm \ref{alg:finding_universal_perturbations} but for VGG-16 (Figure \ref{vgg16}) and VGG-19 (Figure \ref{vgg19}) neural network architecture and on ILSCVR 2012 data.
\end{itemize}

\begin{figure*}[t]
     \centering
     \begin{subfigure}[b]{0.9\textwidth}
         \centering
         \caption{UAP}
         \label{fig:comp1}
         \includegraphics[width=\textwidth]{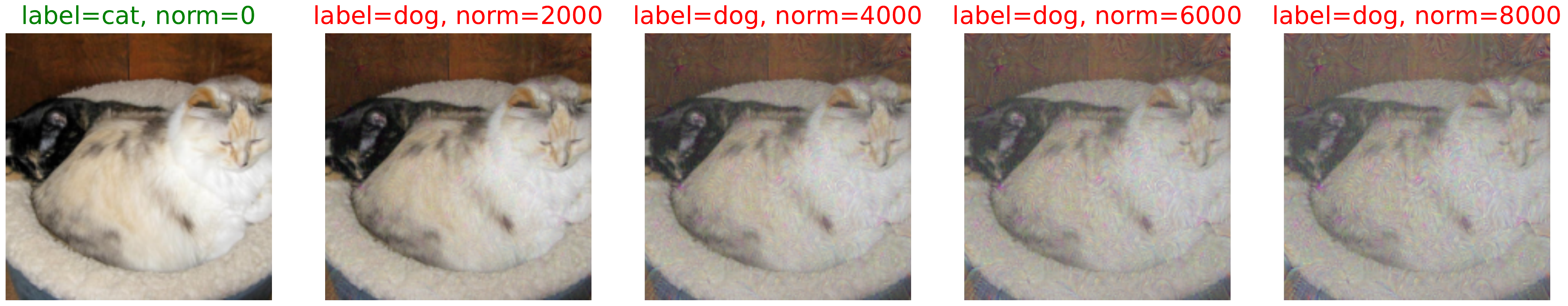}
     \end{subfigure}
     \begin{subfigure}[b]{0.9\textwidth}
         \centering
         \caption{Random}
         \label{fig:comp2}
         \includegraphics[width=\textwidth]{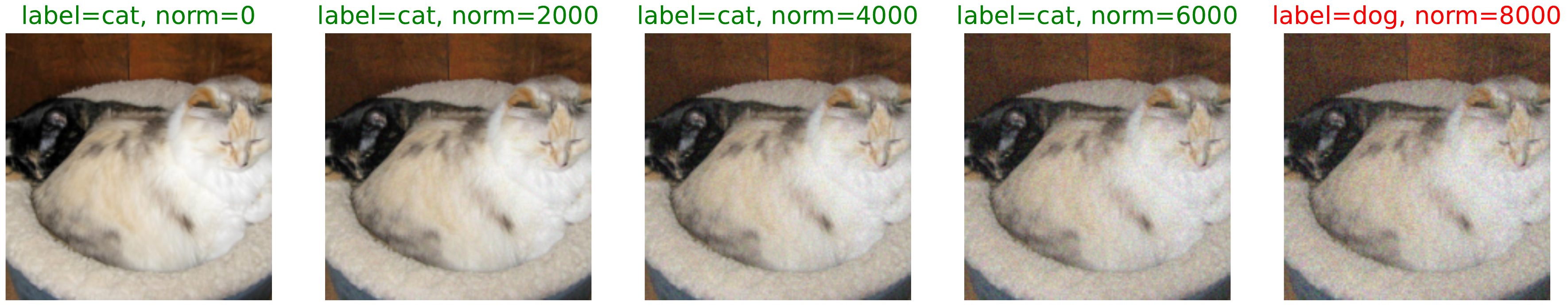}
     \end{subfigure}
        \caption{Evolution of the neural network fooling by varying the perturbation norm}
        \label{fig:comp}
\end{figure*}

First, let's compare our UAP to a random vector. Figure \ref{fig:comp} shows how the label evolves as a relation to the intensity of the perturbation on the image (perturbations are scaled up). Note that on average, a perturbation of $\ell_2$ norm 2000 represents less than 5\% of the pixels in the clean image.

The UAP (Figure \ref{fig:comp1}) manages to fool the neural network on the cat image with a norm of 2,000 while the random vector (Figure \ref{fig:comp2}) must reach a norm of 8,000.

\begin{figure}[h]
\centering
\includegraphics[width=0.45\textwidth]{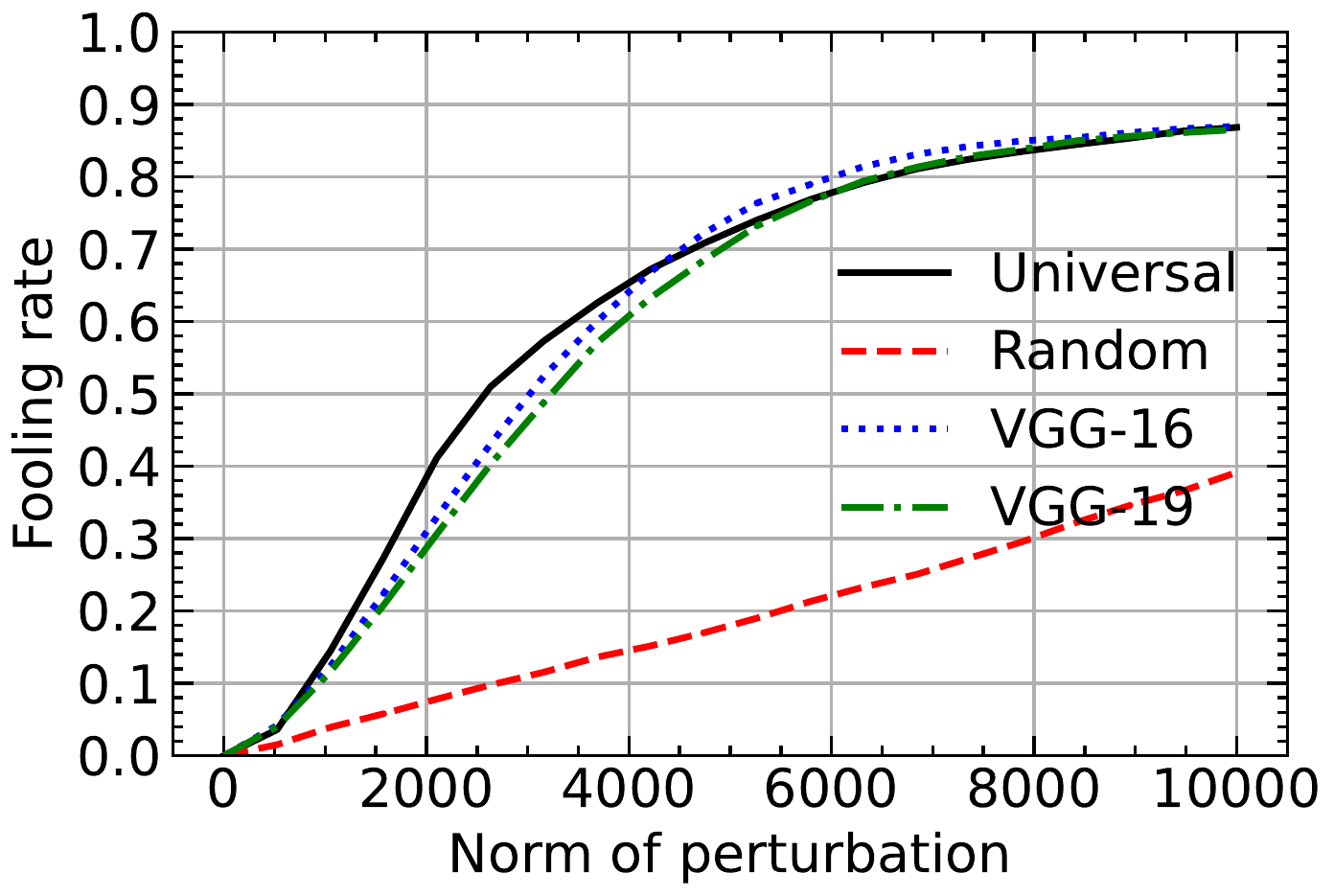}
\caption{Comparison between fooling rates of different perturbations.}
\label{curves} 
\end{figure}

To generalize this observation we display a curve on the graph in Figure \ref{curves} that shows the fooling rate on the validation set of VOC2012 for each perturbation. 

The fact that our UAP (black curve) is more efficient than a random vector seems to be confirmed: it needs a norm of 2000 for the UAP to fool 40\% of the validation set, while it needs a norm of 10,000 for the random vector. The reference paper explains this by suggests that the UAPs exploits some geometric correlations between different parts of the decision boundary of the classifier.

We can also observe the generalization property of UAPs across different neural network architectures. Although the UAPs of VGG-16 and VGG-19 were not trained on the same data and the UAPs are very different, they have very similar results to our UAP on the VOC2012 validation set. This may be a negligible difference, but at around the original norm of our UAP (2,000), our UAP seems to perform slightly better than the other two. I suggest that this is because for the 2,000 norm, our UAP is really very specialised in relation to the data it has learned from. 

We can therefore imagine a modification of Algorithm \ref{alg:finding_universal_perturbations}, to specialise an existing UAP on another architecture, which would consist of taking a pre-calculated vector as input instead of the null vector, to speed up the start of the algorithm and to be able to slightly increase the fooling rate scores. A kind of Transfer Learning applied to UAPs.

\section{Conclusions}

After training a VGG11 neural network on the VOC2012 dataset, we were able to calculate an UAP from the resulting classifier. Despite the size of our dataset, we were able to fool the classifier. We then noticed the existence of dominant labels that contributed much more to the formation of the UAP. Finally we compared our UAP with the random vector and UAPs calculated from other architecture. We saw that our UAP is really more efficient than the random vector, but that it has almost the same performance as the one computed from other neural network architectures (and a much larger dataset).

\section*{\hfil Acknowledgments \hfil}
{\itshape
I thank the authors of this paper, it was a pleasure to work on such an interesting subject; \\
I thank Dr. Michael Clément (Associate professor at Bordeaux Institute of Technology - ENSEIRB-MATMECA School of Engineering) who was a very good supervisor for this work; \\
I thank my friend Hector Piteau for giving me the possibility to access good computational resources for my experiments.
}

\end{document}